%% file: main.tex
\documentclass[runningheads]{llncs}

\usepackage[T1]{fontenc}
\usepackage{graphicx}

\input{preamble}

\begin{document}
\title{WildfireX-SLAM: A Large-scale Low-altitude RGB-D Dataset for Wildfire SLAM and Beyond}
% \author{Zhicong Sun, Jacqueline Lo*, Jinxing Hu*}
\author{Zhicong Sun\inst{1}, Jacqueline Lo*\inst{1}, Jinxing Hu* \inst{2}}

% \authorrunning{}
\institute{Hong Kong Polytechnic University \and Shenzhen Institutes of Advanced Technology, Chinese Academy of Sciences\\ \textasteriskcentered Corresponding authors\\\email{zhicong.sun@connect.polyu.hk, jacqueline.lo@polyu.edu.hk, jinxing.hu@siat.ac.cn}}

% \footnote{ddd}
%
\maketitle              % typeset the header of the contribution
\begin{abstract}
3D Gaussian splatting (3DGS) and its subsequent variants have led to remarkable progress in simultaneous localization and mapping (SLAM). While most recent 3DGS-based SLAM works focus on small-scale indoor scenes, developing 3DGS-based SLAM methods for large-scale forest scenes holds great potential for many real-world applications, especially for wildfire emergency response and forest management. However, this line of research is impeded by the absence of a comprehensive and high-quality dataset, and collecting such a dataset over real-world scenes is costly and technically infeasible. To this end, we have built a large-scale, comprehensive, and high-quality synthetic dataset for SLAM in wildfire and forest environments. Leveraging the Unreal Engine 5 Electric Dreams Environment Sample Project, we developed a pipeline to easily collect aerial and ground views, including ground-truth camera poses and a range of additional data modalities from unmanned aerial vehicle. Our pipeline also provides flexible controls on environmental factors such as light, weather, and types and conditions of wildfire, supporting the need for various tasks covering forest mapping, wildfire emergency response, and beyond. The resulting pilot dataset, WildfireX-SLAM, contains 5.5k low-altitude RGB-D aerial images from a large-scale forest map with a total size of 16 km². On top of WildfireX-SLAM, a thorough benchmark is also conducted, which not only reveals the unique challenges of 3DGS-based SLAM in the forest but also highlights potential improvements for future works. The dataset and code will be publicly available. Project page: \url{https://zhicongsun.github.io/wildfirexslam}.
\keywords{Wildfire  \and Simultaneous Localization and Mapping \and Unmanned Aerial Vehicle.}
\end{abstract}

\section{Introduction}
\label{sec:intro}
Simultaneous Localization and Mapping (SLAM) has continually benefited from advancements in multimedia technologies. Recently, high-fidelity real-time rendering techniques, prominently represented by 3D Gaussian Splatting (3DGS), have significantly propelled the development of SLAM.
Cutting-edge 3DGS-based SLAM, such as SplaTAM~\cite{keetha2023splatam}, MonoGS~\cite{matsuki2023gaussian}, and GS-SLAM~\cite{yan2023gs}, have achieved remarkable success in terms of mapping quality and real-time performance, especially in controlled static indoor environments. Furthermore, methods like DG-SLAM~\cite{yan2023gs} address dynamic tracking challenges by filtering out moving objects, showing state-of-the-art results in dynamic indoor settings. However, despite these developments, the application of 3DGS-based SLAM in large-scale outdoor environments remains underexplored. This gap is particularly prominent in scenarios such as wildfire emergency response and forest management, where robust SLAM systems can play a transformative role.

\begin{figure}[!h]
\includegraphics[width=0.998\linewidth]{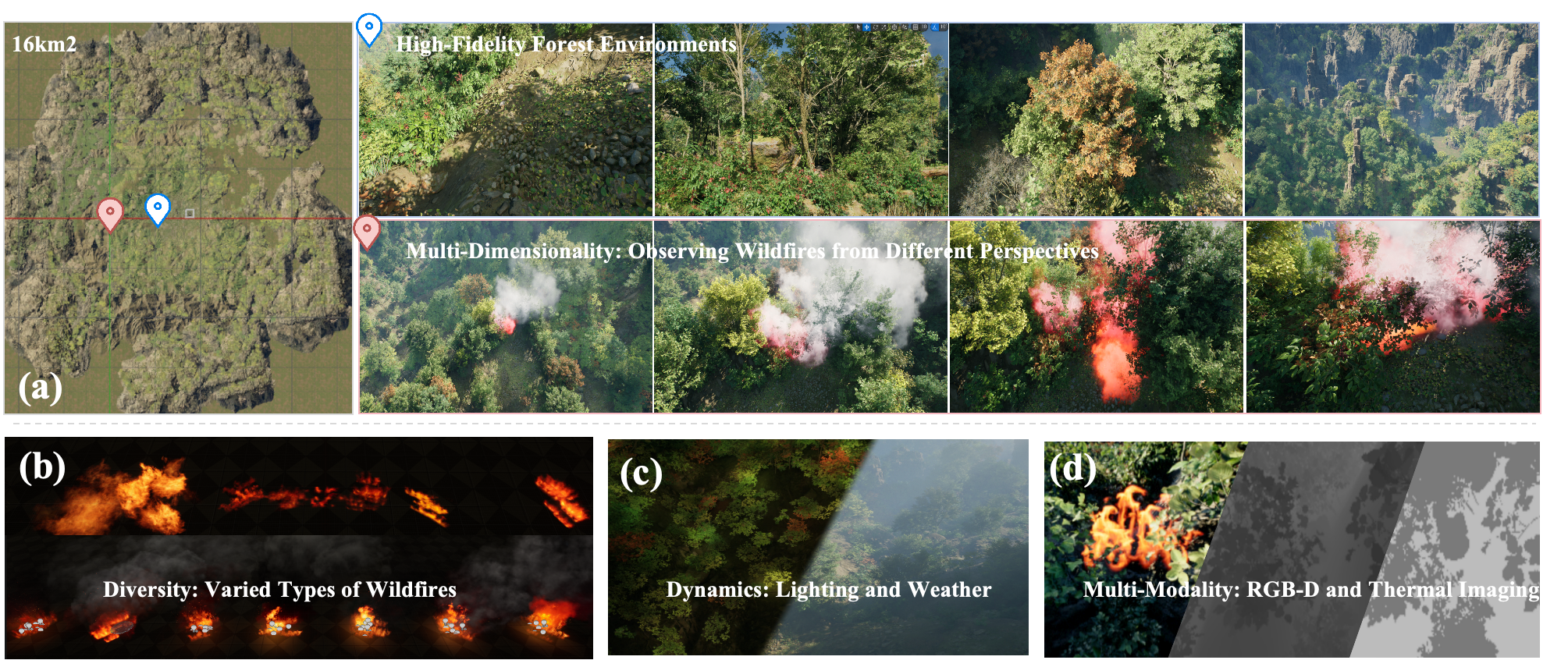}
\caption{Overview of \emph{WildfireX-SLAM}. (a) We have constructed a 16 km² forest scene featuring wildfires, offering a high-fidelity forest environment and multi-dimensional wildfire observations.  (b) Our framework enables the configuration of various types of wildfires. (c) Through diverse weather and lighting conditions, our framework provides high-fidelity and realistic scenes.  (d) We deliver multi-modal data, especially RGB-D from UAVs, to facilitate tasks such as SLAM and navigation.}\label{fig:overview}
\end{figure}

Existing SLAM datasets, such as KITTI~\cite{geiger2012we}, Oxford RobotCar~\cite{maddern20171}, and EuRoC~\cite{burri2016euroc}, have laid the groundwork for advancing SLAM techniques, but they focus on structured, man-made environments. Forest environments present a unique set of challenges due to monotonic textures, sparse structural features, and the dynamic nature of wildfires. Forest-specific datasets, such as SLOAM~\cite{chen2019}, FinnForest~\cite{ali2020finnforest}, and Wild-Places~\cite{knights2023wild}, attempt to address these challenges, but they lack key elements critical for wildfire scenarios, such as dynamic wildfire states, multi-modal observations, and controllable environmental conditions. Furthermore, these datasets typically rely on a single observation platform (either aerial or ground-based), limiting their adaptability to real-world applications that require multi-agent coordination.

To bridge these gaps, we present WildfireX-SLAM, a pioneering SLAM dataset specifically designed for wildfire and forest environments. Built on Unreal Engine 5 and AirSim, this dataset provides synchronized aerial and ground observations, including high-quality RGB, thermal, and depth imagery, accompanied by ground-truth camera poses. WildfireX-SLAM enables realistic simulations with controllable environmental factors, such as wildfire types, weather, and lighting conditions, offering a flexible platform for diverse SLAM-related tasks, including wildfire emergency response, forest mapping, and swarm-based perception. With 5.5k aerial images collected over a 16 km² forest map, WildfireX-SLAM represents a comprehensive resource for SLAM research in these challenging environments.

In addition to introducing the dataset, we perform extensive benchmarking on WildfireX-SLAM, revealing critical challenges and unique behaviors exhibited by 3DGS-based SLAM methods in wildfire and forest scenarios. These experiments not only expose the limitations of current techniques but also highlight opportunities for developing robust methods tailored to these environments. We believe WildfireX-SLAM will serve as a valuable tool for the community, accelerating research on emergency response, environmental monitoring, and beyond.

In summary, our contributions are as follows:
\begin{enumerate}
    \item \textbf{Development of the WildfireX-SLAM Dataset:} We present WildfireX-SLAM, a large-scale and high-quality SLAM dataset specifically designed for wildfire environments. This dataset highlights key attributes of wildfire scenarios, including complex dynamic forest environments, diverse wildfire types, and variable weather conditions. WildfireX-SLAM consists of aerial imagery captured by UAVs flying at low altitudes below 50 meters above ground, supplemented with additional data modalities such as depth, supporting a wide range of SLAM-related tasks.
    \item \textbf{Creation of a Flexible Data Collection Framework:} We developed a robust framework leveraging Unreal Engine 5 and AirSim to automate the collection of high-quality forest data. This framework allows researchers to flexibly control wildfire states, weather conditions, and sensor configurations, simplifying data acquisition for various task setups. It is a valuable community resource, enabling users to generate advanced datasets tailored to their specific research needs.
    \item \textbf{Extensive Benchmarking and Insights:} We conducted comprehensive experiments on the WildfireX-SLAM dataset to uncover key challenges in performing SLAM in wildfire and forest environments. These insights aim to drive future research efforts and advancements in this domain.
\end{enumerate}

\section{Related Works}
\label{sec:related_work}

Recently, benefiting from the explicit map representation and real-time high-fidelity rendering capabilities provided by 3D Gaussian Splatting (3DGS)~\cite{kerbl20233d}, SLAM technology has made significant advancements in both real-time performance and mapping quality, bringing it closer to practical deployment. 
3DGS-based methods, such as SplaTAM~\cite{keetha2023splatam}, MonoGS~\cite{matsuki2023gaussian}, and GS-SLAM~\cite{yan2023gs}, have achieved state-of-the-art (SOTA) performance in terms of map reconstruction quality and speed within static environments. 
DG-SLAM~\cite{yan2023gs} attempts to address tracking challenges in dynamic environments by filtering out dynamic objects, achieving SOTA performance in dynamic indoor scenes. 
However, most of these approaches remain confined to controlled indoor scenarios, with limited exploration of their application in outdoor emergency response contexts, where SLAM could play a crucial role. The primary underlying reason for this limitation lies in the absence of relevant datasets.

In the field of SLAM, numerous representative datasets, such as MIT-DARPA~\cite{huang2010high}, Rawseeds~\cite{fontana2014rawseeds}, KITTI~\cite{geiger2012we}, NCLT~\cite{carlevaris2016university}, Oxford RobotCar~\cite{maddern20171}, KAIST Day/Night~\cite{wenzel20214seasons}, 4-Seasons~\cite{choi2018kaist}, ComplexUrban~\cite{jeong2019complex}, UrbanLoco~\cite{wen2020urbanloco}, TUM-RGBD~\cite{sturm2012benchmark}, EuRoC~\cite{burri2016euroc}, TUM-VI~\cite{schubert2018tum}, OpenLORIS~\cite{shi2020we}, M2DGR~\cite{yin2021m2dgr}, Newer College~\cite{ramezani2020newer}, and Hilti SLAM~\cite{helmberger2022hilti}, have been developed. These datasets cover a wide range of real-world challenges in structured environments, providing invaluable opportunities for algorithm validation and improvement.
However, emergency SLAM applications often operate in unstructured environments that lack clear, regular, or well-defined features. Forest environments exemplify such conditions, presenting monotonic textures and limited structural cues, which pose a series of challenges to SLAM systems. Currently, the following datasets focus on forest-specific scenarios:

SLOAM~\cite{chen2019} dataset utilizes LiDAR scans in dense forest environments to enable semantic-aware SLAM, providing ground-truth labels for terrain and tree instances. It addresses vegetation occlusion and sparse geometric features, supporting applications in forest inventory and autonomous navigation under unstructured conditions.

% \subsection{Datasets and Benchmarks for SLAM in Forest}
FinnForest~\cite{ali2020finnforest} and FinnWoodlands~\cite{FinnWoodlands} datasets can be jointly used to address forest environment perception. FinnForest captures multi-sensor seasonal data (RGB, IMU, GNSS) for evaluating SLAM robustness under natural lighting and terrain dynamics, while FinnWoodlands provides static scenes with stereo RGB, LiDAR point clouds, and semantic instance labels (tree species, terrain types) to advance fine-grained environmental understanding and depth completion models.

Wild-Places~\cite{knights2023wild} dataset is a longitudinal LiDAR-based collection designed to evaluate long-term autonomy in unstructured natural environments. It comprises 63k georeferenced submaps captured over 14 months across seasonal variations (e.g., vegetation growth/decay), addressing challenges in place recognition and semantic segmentation under GNSS-denied conditions. With multi-temporal revisits and domain shifts induced by natural environmental dynamics, it supports robustness analysis of lidar odometry and retrieval systems in dense forestry scenarios.

BotanicGarden~\cite{liu2024botanicgarden} dataset provides multi-sensor trajectories (spinning/MEMS LiDARs, stereo cameras, IMUs) spanning 17.1 km in a 48k m² botanical garden, focusing on SLAM performance in texture-poor, slope-varied terrains. It features synchronized ground-truth maps, annotated semantic obstacles, and traversability labels, specifically targeting navigation challenges in vegetation-dense, GNSS-deprived ecosystems with weak visual features and complex geometric structures.

Although the aforementioned datasets have made significant contributions to SLAM in forest environments, their lack of wildfire-specific elements renders them unsuitable for SLAM tasks in wildfire scenarios. Furthermore, these datasets rely on a single observation platform, either ground-based vision or aerial perspectives, and their lighting and weather conditions are entirely dependent on the time of data collection, making them uncontrollable.
The statistics and properties of these related datasets are summarized in \Cref{tab:dataset_comparison}.

In contrast, the WildfireX-SLAM dataset proposed in this paper is the first specifically designed for wildfire observation. Beyond incorporating wildfire elements, our dataset provides more comprehensive observation perspectives, including both near-ground and aerial views. Moreover, lighting and weather conditions are fully controllable, which enables the realistic simulation of the challenges that SLAM and navigation tasks face in wildfire emergency response scenarios.
Our dataset can also be extended to support multi-agent swarm-based emergency environment perception tasks. This scalability allows for the simulation and study of coordinated operations involving multiple UAVs and unmanned ground vehicles in dynamic wildfire scenarios, further enhancing its applicability to real-world emergency response and collaborative SLAM research.

\begin{table}[!h]
    \caption{Comparison of statistics and properties between our {\em WildfireX-SLAM} dataset and related datasets.}
    \label{tab:dataset_comparison}
\small
  \centering
  \resizebox{\linewidth}{!}{
    \begin{tabular}{cccccccccc}
    \toprule
    Dataset & \multicolumn{1}{l}{Data} & Scene & Viewpoint & Source & Lighting & Weather & Forest/Tree & Flame/Smoke \\ \midrule
    S3E~\cite{feng2024s3e} & \multicolumn{1}{l}{RGB-D \& Heat \& Lidar} & Campus & Ground & Real & \xmark & \xmark & \checkmark & \xmark \\
    KITTI 2~\cite{gaidon2016virtual} & \multicolumn{1}{l}{RGB-D} & City & Ground & Synthetic \& Real & \checkmark & \checkmark & \xmark & \xmark \\
    BASEPROD~\cite{gerdes2024baseprod} & \multicolumn{1}{l}{RGB-D \& Heat} & Desert & Ground & Real & \xmark & \xmark & \xmark & \xmark \\ 
    SLOAM~\cite{chen2019} & \multicolumn{1}{l}{Lidar} & Forest & Aerial & Real & \xmark & \xmark & \checkmark & \xmark \\
    FinnForest~\cite{ali2020finnforest} & \multicolumn{1}{l}{RGB} & Forest & Ground & Real & \checkmark & \checkmark & \checkmark & \xmark \\ 
    FinnWoodlands~\cite{FinnWoodlands} & \multicolumn{1}{l}{RGB-D \& Lidar} & Forest & Ground & Real & \checkmark & \checkmark & \checkmark & \xmark \\ 
    Wild-places~\cite{knights2023wild} & \multicolumn{1}{l}{Lidar} & Forest & Ground & Real & \checkmark & \checkmark & \checkmark & \xmark \\ 
    Botanicgarden~\cite{liu2024botanicgarden} & \multicolumn{1}{l}{RGB \& Lidar} & Forest & Ground & Real & \checkmark & \checkmark & \checkmark & \xmark \\ 
    \bottomrule\midrule
    \textbf{Ours} & \multicolumn{1}{l}{RGB-D \& Heat} & Forest & Aerial+Near-ground & Synthetic & \checkmark & \checkmark & \checkmark & \checkmark \\ \bottomrule
    \end{tabular}%
    }
\end{table}%
\vspace{-3em}
\section{WildfireX-SLAM Dataset}
\label{sec:method}
The WildfireX-SLAM dataset aims to establish an unprecedented and challenging benchmark for SLAM and navigation in the context of wildfire emergency response by providing comprehensive wildfire scene maps observed from UAVs.
In addition to RGB images, we provide depth images, normal maps, and thermal images observed by sensors on UAVs. Furthermore, we include their motion data, such as current position, velocity, acceleration, and camera poses, to support various other tasks.
Furthermore, we offer flexible control over environmental factors, such as forest complexity, the types, colors, brightness, and spread rates of smoke and fire elements, as well as the direction and intensity of light and the density of fog, to simulate complex and dynamic real-world scenarios. \Cref{sec:3.1} details the data construction process, while \Cref{sec:3.3} present comprehensive statistics and characteristics of the dataset.

\subsection{Dataset Construction}
\label{sec:3.1}

\textbf{Construction of Wildfire Scenarios.} Due to the limitations of human resources and the priority of emergency rescue, deploying swarms of UAVs and unmanned vehicles to collect sufficient data in real-world wildfire scenarios is impractical. Therefore, we constructed high-fidelity wildfire scenarios using Unreal Engine 5.
The rationale behind using UE and Nigara to simulate wildfire as well as the effectiveness of sim-to-real transfer have been discussed and shown in the FireFly dataset~\cite{hu2023firefly}.
Our synthetic wildfire scene seamlessly integrates two originally independent Unreal Engine 5 systems: the Electric Dreams Environment Sample Project, used for procedural biome generation with PCG-controlled terrain and Quixel Megascans assets, and the M5 VFX Niagara Fire System, which enables Niagara-based particle simulations with 4096×4096 fire and smoke textures. By embedding these systems into a unified UE5 project, we synchronize environmental parameters via Unreal Blueprints, dynamically linking PCG-generated variables such as vegetation density and slope gradients to the combustion properties of the M5 VFX system, including flame spread velocity and smoke particle density.
This integration leverages Unreal Engine 5’s native interoperability, allowing the creation of wildfire scenarios in which fire propagation adapts to procedurally generated fuel distributions.
The constructed wildfire scenario, as illustrated in \Cref{fig:overview} (a), encompasses a 16-square-kilometer forest environment featuring diverse forest landscapes. The wildfires can be observed from any perspectives and at different altitudes.

\begin{figure}[!h]
\includegraphics[width=0.998\linewidth]{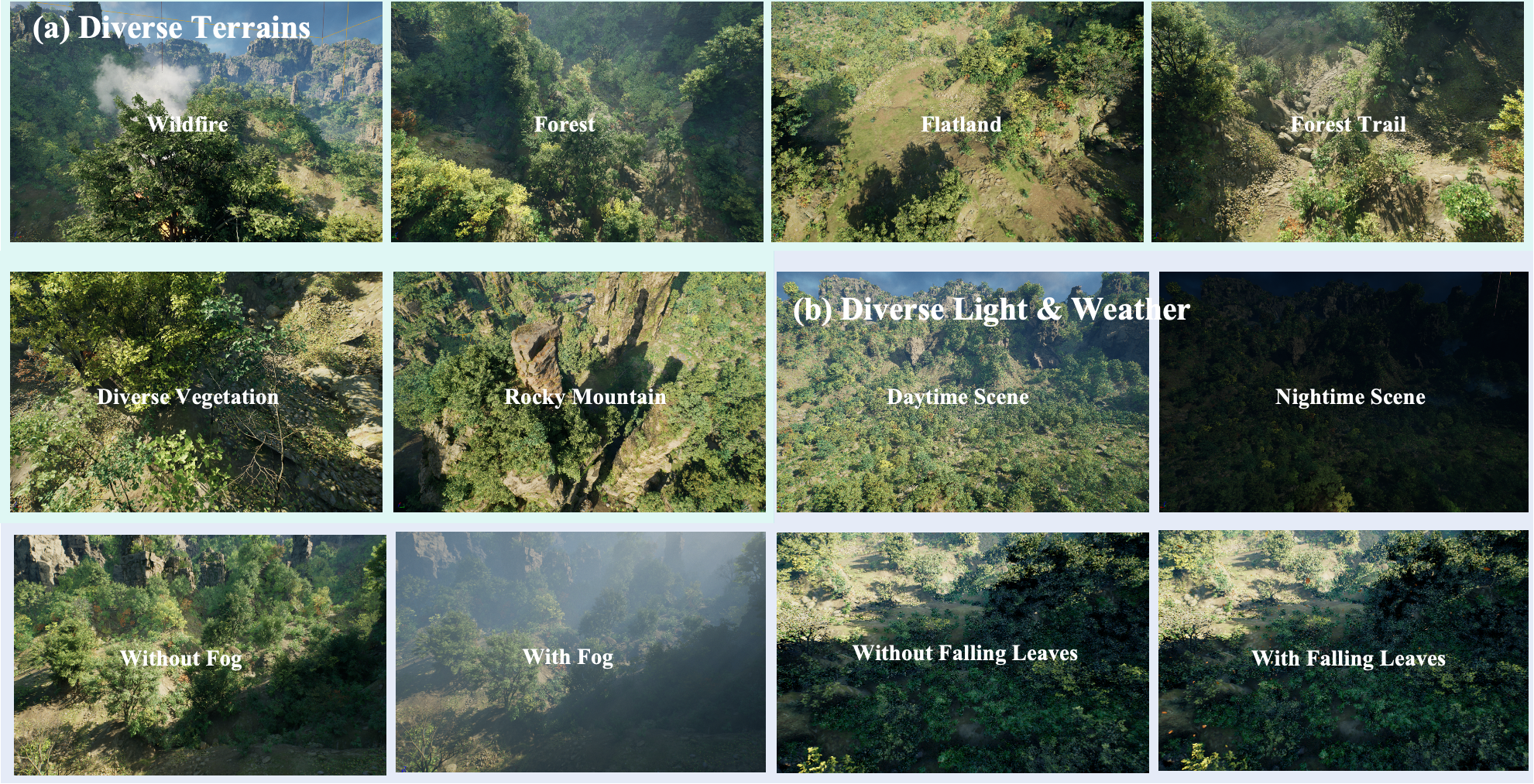}
\caption{Diversity of the \emph{WildfireX-SLAM} dataset.}\label{fig:diversity}
\end{figure}
\textbf{Data Collection Tool.} SLAM and navigation in wildfire scenarios using UAVs require the dense collection of 2D images, depth maps, and thermal images to learn accurate scene representations and to avoid hazardous areas. In Unreal Engine 5, collecting such data is a complex process that involves setting up adjustable camera parameters to capture specific viewpoints. Although Unreal Engine 5 provides the Movie Render Queue plugin for high-quality image rendering, manually specifying keypoint positions, rotations, and frame numbers can be both time-consuming and inflexible. To address this challenge, we modified AirSim, an open-source simulator developed by Microsoft, and customized its code for our specific recording requirements. Our modified AirSim provides a high-fidelity simulation platform for aerial and ground vehicles, integrating realistic physics and environmental dynamics. It is equipped with APIs for automating data collection, supporting the generation of required sensor data. By incorporating AirSim, we significantly reduced the need for manual annotations and improved the efficiency of our data collection process.

\begin{figure}[!h]
\begin{center}
\includegraphics[width=1.0\textwidth]{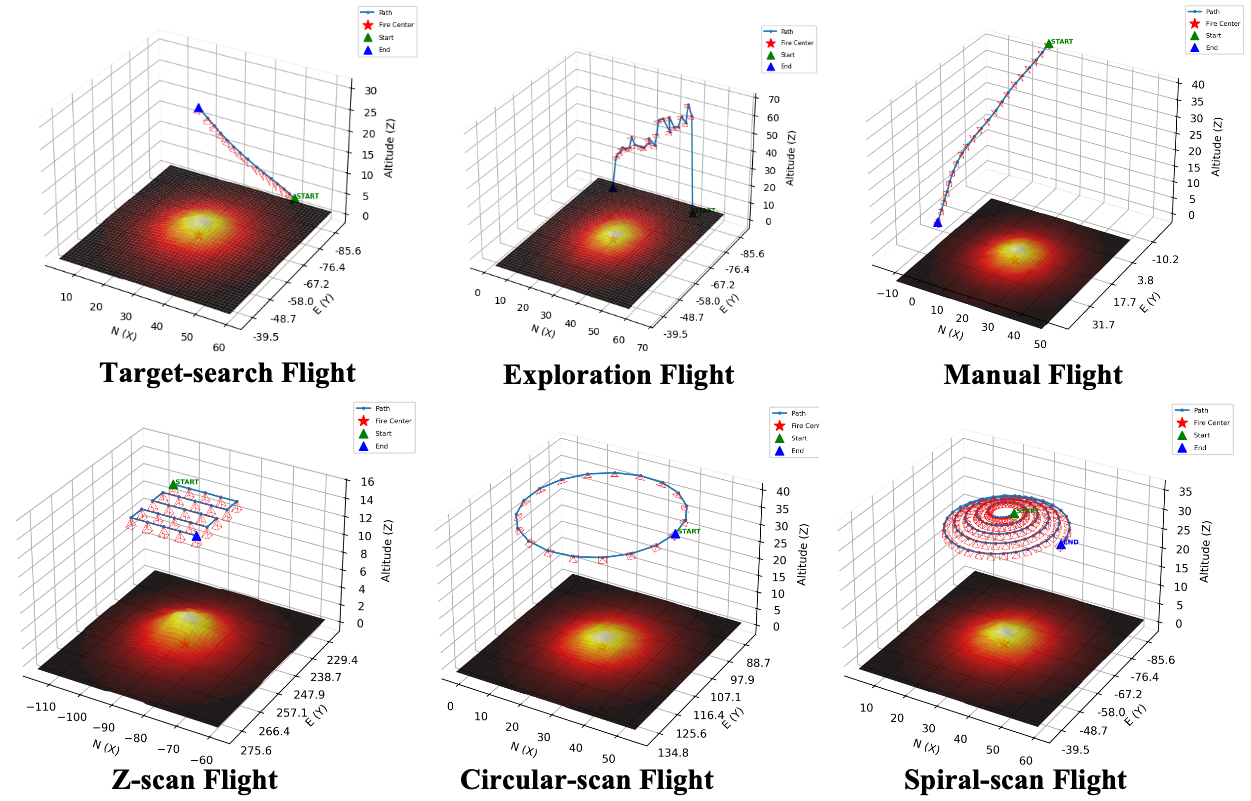}
\end{center}
% \vspace{-1em}
   \caption{Examples of flight path for data collection in {\em WildfireX-SLAM} dataset.}
\label{fig:3dpath}
\end{figure}

\textbf{Data Collection Using UAVs.}
During the data collection, UAVs are deployed to collect multi-modal data through predefined or manual flight paths, all executed within the same dynamic environment. This ensures that the dataset can simulate the concurrent operation of multiple UAVs, supporting research on swarm-based wildfire perception. Each UAV is equipped with a RGB-D camera, both mounted centrally underneath the UAV body, with their orientations can be changed in real-time according to the flight modes.
Considering the complexity and diversity of flight patterns in wildfire environments, the data collection encompasses exploratory flights during periods when the wildfire locations are unknown, as well as target-search flights, manual flights, and scanning flights such as Z-scan, spiral-scan, and circular-scan after the wildfire areas have been identified, as illustrated in \Cref{fig:3dpath}.
The flight speed is maintained at 3 meters per second, with obstacle avoidance dynamically adjusting the path. All data, including RGB-D images and motion parameters (\eg, position, velocity, orientation), is recorded at 30 Hz to support dynamic wildfire mapping and SLAM research.

\subsection{Dataset Characteristics}
\label{sec:3.3}

\textbf{Large-Scale, Diversity, and Low-altitude.} The WildfireX-SLAM dataset spans a 16 square kilometer forest environment, a significant scale compared to most existing datasets. It includes 5.5k aerial images captured by UAVs, ensuring comprehensive coverage of the dynamic wildfire scenario. Unlike other datasets~\cite{ali2020finnforest,FinnWoodlands,knights2023wild,liu2024botanicgarden}, which focus on isolated and small-scale scenes, WildfireX-SLAM provides a densely captured, unified forest environment from various viewpoints, as shown in \Cref{fig:diversity}. Furthermore, unlike datasets such as S3E~\cite{feng2024s3e} and KITTI 2~\cite{gaidon2016virtual}, which limits coverage to urban region, or BASEPROD~\cite{gerdes2024baseprod}, which does not feature complete forest contexts, WildfireX-SLAM enables robust analysis of forest-specific SLAM and navigation.
Additionally, unlike other wildfire datasets collected from UAVs flying at high altitudes~\cite{hu2023firefly,li2024sim2real}, our dataset focuses on low-altitude flights below 50 meters above ground. This relatively rare but feasible flying altitude provides richer environmental details, which are crucial for wildfire perception and emergency response.

\textbf{Controllable Environments.} One of WildfireX-SLAM’s unique strengths lies in its ability to simulate dynamic, controllable environments. Researchers can manipulate key environmental factors, including flame and smoke dynamics, lighting angles and intensities, and fog density and height, as demonstrated in \Cref{fig:overview} (b) and (c). These controls allow for realistic simulation of complex wildfire conditions, supporting experiments with diverse scene configurations that would be challenging to capture in real-world data. Moreover, Unreal Engine facilitates the emulation of camera noise, such as motion blur and defocus blur, further enhancing realism. This flexibility ensures that WildfireX-SLAM bridges the gap between synthetic data and real-world applications.

\textbf{Multiple Attributes.} As demonstrated in \Cref{fig:overview} (d), Aprting from RGB-D data, the WildfireX-SLAM can also provide a wide range of intermediate data products, including thermal images, normals, and decomposed reflectance components. These attributes are critical for advanced tasks such as SLAM, depth estimation, and navigation research. Thanks to our integrated pipeline, these properties can be extracted at minimal additional computational cost, eliminating the prohibitive expenses associated with acquiring such data in real-world scenarios.

\textbf{Applications.} By exploring SLAM and navigation algorithms on WildfireX-SLAM, researchers can bridge the gap between synthetic and real-world applications. The dataset’s high-quality imagery, precise camera poses, dynamic wildfire scenarios, and diverse sensor attributes make it invaluable for wildfire monitoring, collaborative navigation, and other research areas. Furthermore, its comprehensive design facilitates studies on forest-specific challenges, paving the way for effective real-world deployment.

\section{Experiments}
\label{sec:experiments}
In this section, we primarily investigate the accuracy of camera pose estimation as well as the quality of map reconstruction and rendering. Furthermore, we identify and highlight the challenges faced by representative 3DGS-based methods when applied to this task.

\subsection{Dataset and Metrics}
\textbf{WildfireX-SLAM benchmark.}
WildfireX-SLAM dataset is divided into two benchmark subsets. Benchmark A consists of two sequences of UAV low-altitude flights (within 15 meters above ground), where Sequence 2 contains wildfire presence and Sequence 1 does not. This benchmark is designed to compare the impact of wildfires on various SLAM algorithms. Benchmark B is a more comprehensive benchmark intended to assess algorithm performance across four challenging wildfire scenarios. Challenge Scenario 1 (CS 1) evaluates the effects of dynamic non-rigid objects in the environment, such as wildfires and falling leaves, on SLAM; it includes four sequences: the first two sequences are collected in the same environment without and with  wildfire, respectively, while the latter two sequences are from another wildfire environment with one sequence lacking leaves and the other containing leaves. Challenge Scenario 2 (CS 2) examines the impact of lighting conditions on wildfire SLAM, comprising three sequences collected at the same scene during morning, noon, and late night. Challenge Scenario 3 (CS 3) assesses the influence of weather conditions, containing two sequences—one without fog and one with fog. Challenge Scenario 4 (CS 4) integrates all adverse environmental factors, including wildfire, falling leaves, fog, coupled with poor lighting conditions during late night.

\textbf{Metrics.}
To evaluate the tracking and mapping performance of each baseline on our dataset, we adopt widely used metrics for camera tracking accuracy and map quality as in \cite{Sandström2023ICCV}. For camera tracking, we report the Root Mean Square Error (RMSE) of the Absolute Trajectory Error (ATE) in meters. For map quality, we report photometric rendering metrics, including Peak Signal-to-Noise Ratio (PSNR), Structural Similarity Index (SSIM), and Learned Perceptual Image Patch Similarity (LPIPS).

\subsection{Baselines}
We aim to evaluate the performance of current 3DGS-based SLAM methods on the WildfireX-SLAM dataset to explore the challenges posed by dynamic wildfire emergency scenarios. The decision not to select other SLAM methods designed for dynamic environments is based on the fact that 3DGS-based SLAM currently achieves SOTA performance in both mapping quality and speed, making it particularly well-suited for emergency wildfire scenarios that demand high real-time capabilities. Moreover, its potential in dynamic environments remains limited exploration. Although these methods currently rely on substantial memory resources, this limitation is acceptable and is expected to be addressed in the future. For our experiments, we selected two of the most representative 3DGS-based methods MonoGS and SplatAM. In all experiments, we utilized the official implementations of these baseline methods.

\subsection{Results and Discussion}
\Cref{tab:benchmarka} presents the performance of different algorithms under wildfire and non-wildfire conditions. The presence of fire significantly degrades the localization and mapping performance of the tested algorithms. Specifically, PSNR nearly drops by half, while ATE RMSE increases by more than double. A comparative analysis across the algorithms reveals that the 3DGS-based methods demonstrate localization capabilities comparable to the traditional ORB-SLAM3. Concurrently, their mapping performance surpasses that of the NeRF-based Co-SLAM. Considering the image rendering speed, 3DGS-based SLAM remains the most promising research direction for wildfire scenarios.

\begin{table}[!h]
\caption{Evaluations of MonoGS, Splatam, Co-SLAM and ORB-SLAM3 on Benchmark A of the \emph{WildfireX-SLAM} dataset.}
\label{tab:benchmarka}
\centering
\resizebox{\linewidth}{!}{
\begin{tabular}{c|ccccc|ccccc|ccccc|c}
\toprule
& \multicolumn{5}{c|}{\textbf{MonoGS~\cite{matsuki2023gaussian}}} 
& \multicolumn{5}{c|}{\textbf{Splatam~\cite{keetha2023splatam}}}
& \multicolumn{5}{c|}{\textbf{Co-SLAM}} 
& \multicolumn{1}{c}{\textbf{ORB-SLAM3}} \\
\multirow{-2}{*}{Scene} 
& ATE & PSNR $\uparrow$ & SSIM $\uparrow$ & LPIPS $\downarrow$ & Depth L1 $\downarrow$ 
& ATE & PSNR $\uparrow$ & SSIM $\uparrow$ & LPIPS $\downarrow$ & Depth L1 $\downarrow$ 
& ATE $\downarrow$ & PSNR $\uparrow$ & SSIM $\uparrow$ & LPIPS $\downarrow$ & Depth L1 $\downarrow$ 
& ATE $\downarrow$ \\
\midrule
Sequence 1
& 2.5 & 22.94 & 0.622 & 0.455 & 1.07
& 3.1 & 22.23 & 0.652 & 0.428 & 0.83
& 6.30 & 21.54 & 0.702 & 0.353 & 1.44
& 2.10 \\
Sequence 2
& 12.5 & 13.47 & 0.562 & 0.649 & 6.13
& 13.4 & 13.34 & 0.551 & 0.549 & 4.70
& 52.8 & 6.77 & 0.412 & 0.635 & 7.60
& 9.24 \\
% \midrule
Overall 
& 7.50 & 18.21 & 0.592 & 0.552 & 3.60
& 8.25 & 17.79 & 0.602 & 0.488 & 2.77
& 29.55 & 14.16 & 0.557 & 0.494 & 4.52
& 5.67 \\
\bottomrule
\end{tabular}
}
\end{table}
\vspace{-2em} 
\begin{table}[!h]
\caption{Evaluations of 3DGS-based SLAM on Benchmark B of the \emph{WildfireX-SLAM} dataset.}
\label{tab:benchmarkb}
\small
\centering
\resizebox{\linewidth}{!}{
\begin{tabular}{c|cc|cc|cc|ccc|cc|c}
\toprule
\multirow{2}{*}{Metrics} & \multicolumn{4}{c|}{CS 1} & \multicolumn{2}{c|}{CS 2} & \multicolumn{3}{c|}{CS 3} & \multicolumn{2}{c|}{CS 4} & \multirow{2}{*}{CS 5} \\
 & w/o fire & w fire & w/o leaf & w leaf & w/o & w & morn & noon & night & w/o & w & \\
\midrule
ATE RMSE $\downarrow$& 95.2213 & 98.2351 & 148.8109 & 154.2466 & 51.5609 & 59.618 & 98.4574 & 92.8674 & 92.1550 & 44.8714 & 59.7346 & 108.0769 \\
PSNR $\uparrow$ & 15.60 & 14.77 & 17.96 & 18.02 & 17.50 & 16.70 & 14.04 & 15.12 & 17.81 & 42.06 & 51.94 & 44.74\\
Depth RMSE $\downarrow$ & 12.1128 & 14.8473 & 4.9928 & 6.7917 & 7.7263 & 9.9687 & 20.5924 & 12.0614 & 11.2009 & 28.6708 & 24.9421 & 17.4935\\
MS-SSIM $\uparrow$ & 0.614 & 0.609 & 0.789 & 0.705 & 0.700 & 0.670 & 0.575 & 0.650 & 0.771 & 0.763 & 0.870 & 0.832 \\
LPIPS $\downarrow$ & 0.652 & 0.664 & 0.600 & 0.673 & 0.514 & 0.673 & 0.725 & 0.642 & 0.509 & 0.326 & 0.200 & 0.351 \\
\bottomrule
\end{tabular}}
\end{table}

\begin{figure}[!h]
\begin{center}
\includegraphics[width=1.0\textwidth]{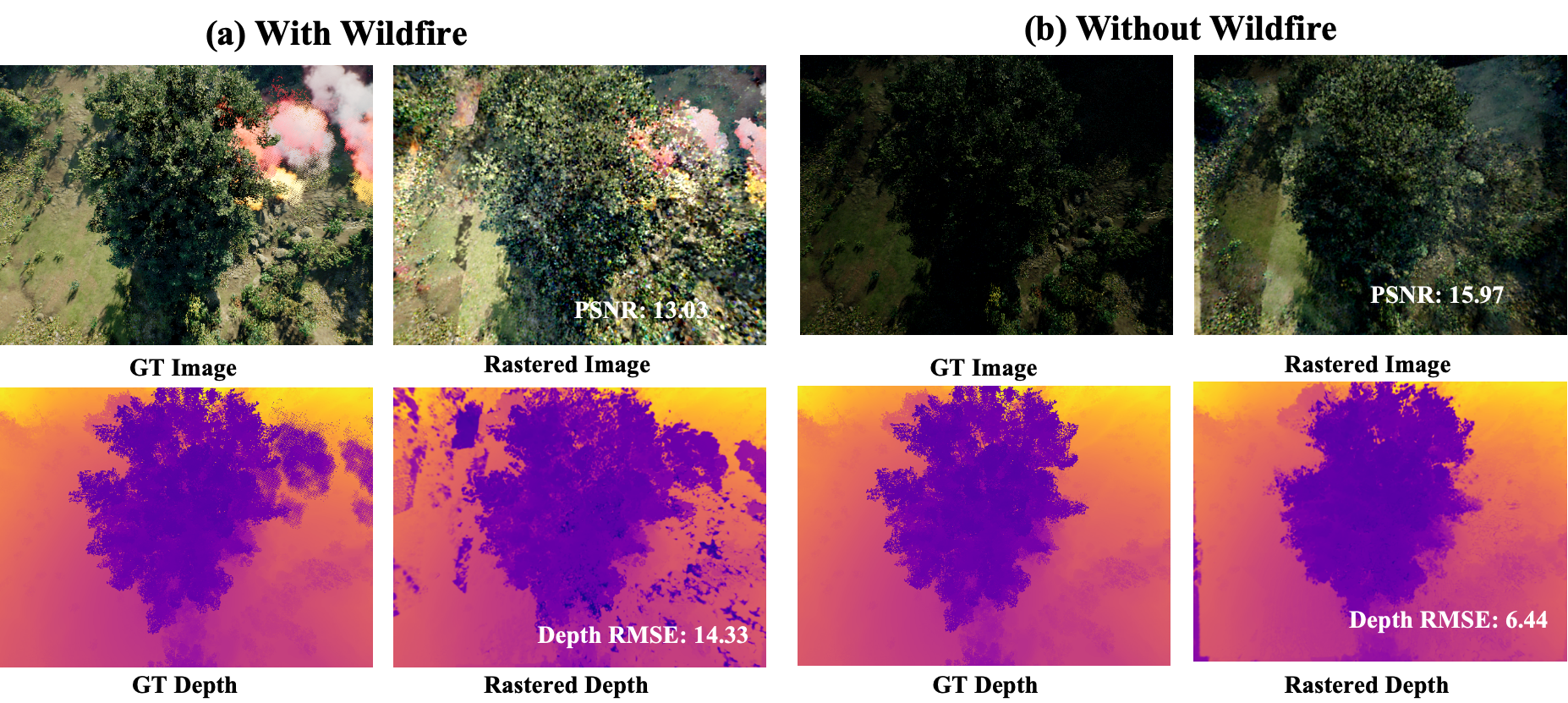}
\end{center}
   \caption{Mapping results of 3DGS-based SLAM on Challenge Scenario 1 of the {\em WildfireX-SLAM} dataset.}
\label{fig:CS1}
\end{figure}
\begin{figure}[!h]
\begin{center}
\includegraphics[width=1.0\textwidth]{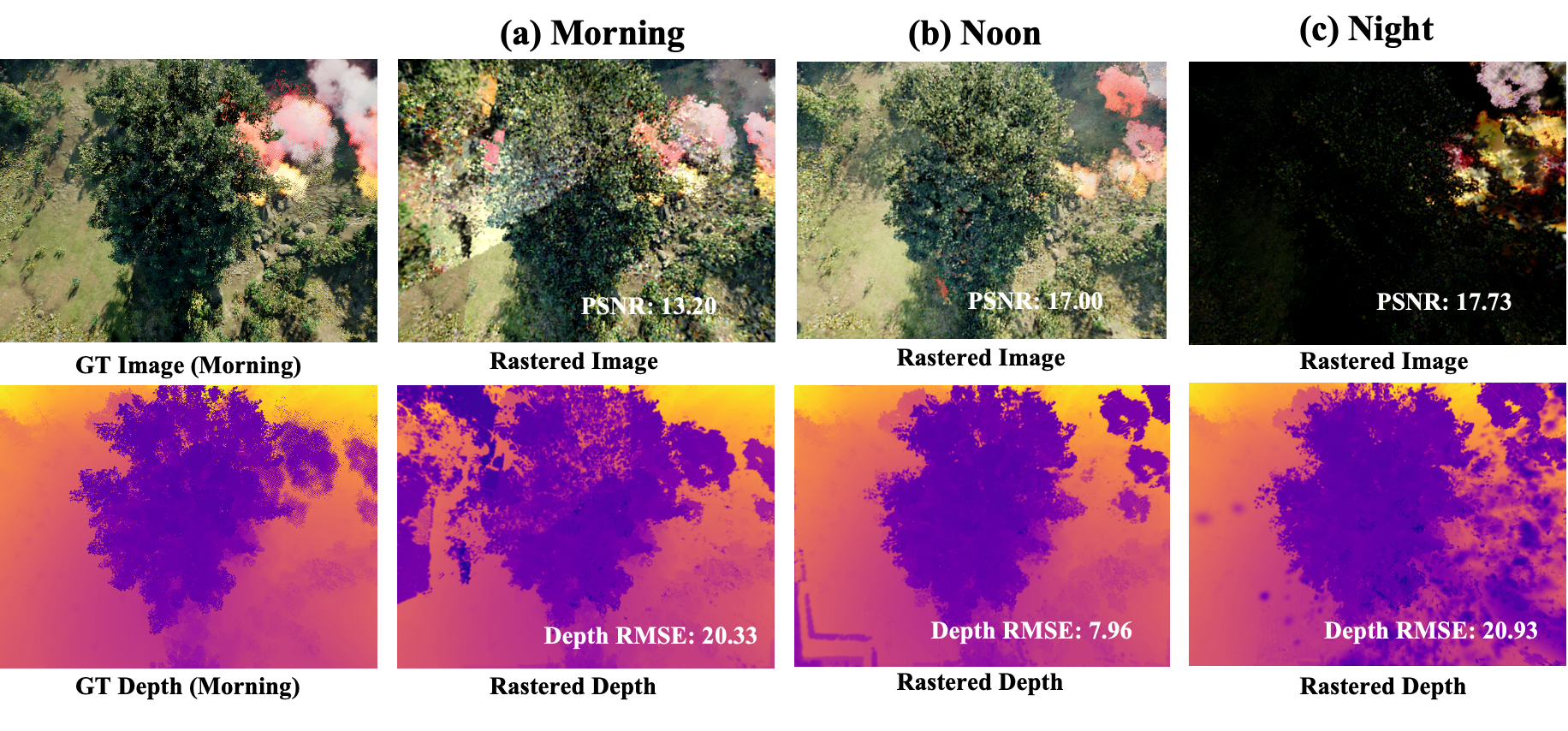}
\end{center}
   \caption{Mapping results of 3DGS-based SLAM on Challenge Scenario 2 of the {\em WildfireX-SLAM} dataset.}
\label{fig:CS2}
\end{figure}
\begin{figure}[!h]
\begin{center}
\includegraphics[width=1.0\textwidth]{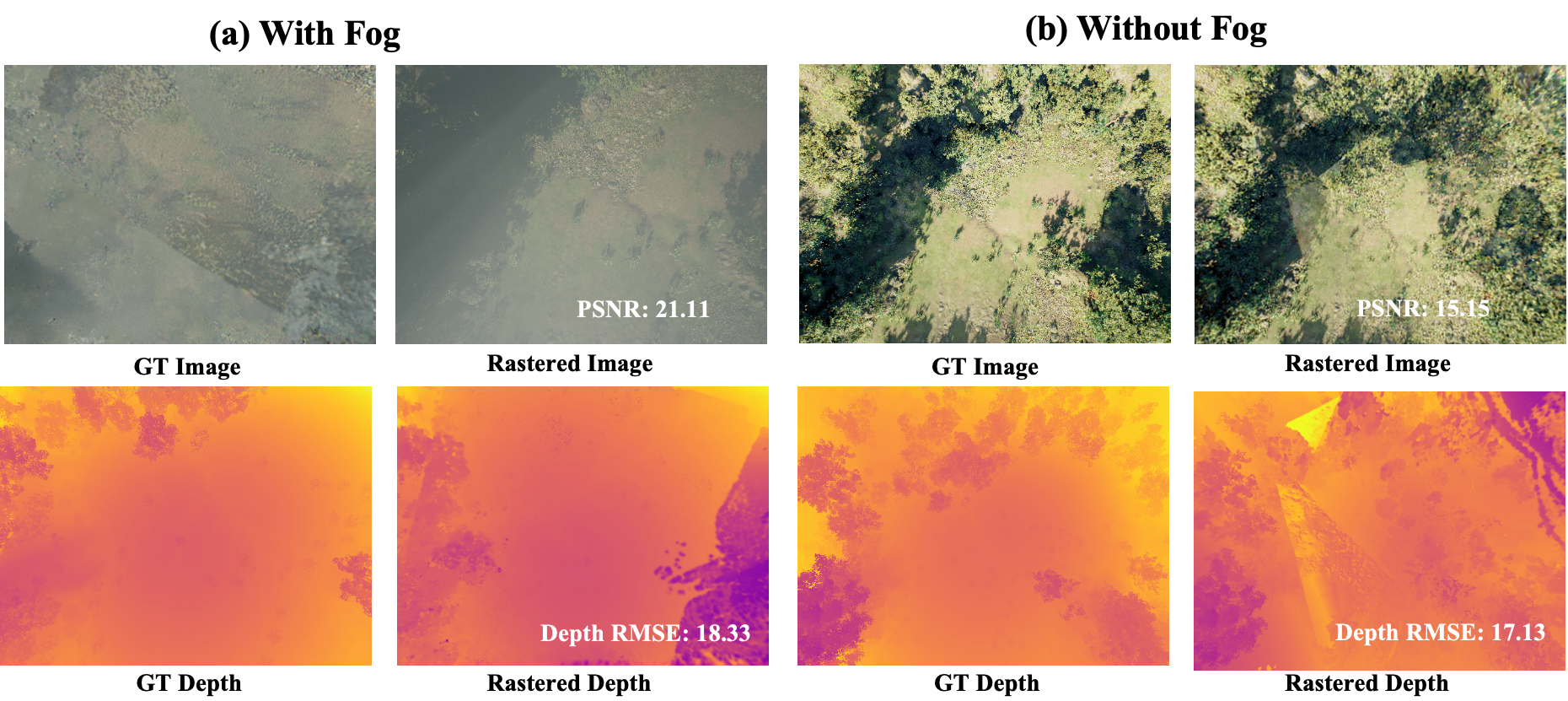}
\end{center}
% \vspace{-1em}
   \caption{Mapping results of 3DGS-based SLAM on Challenge Scenario 3 of the {\em WildfireX-SLAM} dataset.}
\label{fig:CS3}
\end{figure}
\begin{figure}[!h]
\begin{center}
\includegraphics[width=1.0\textwidth]{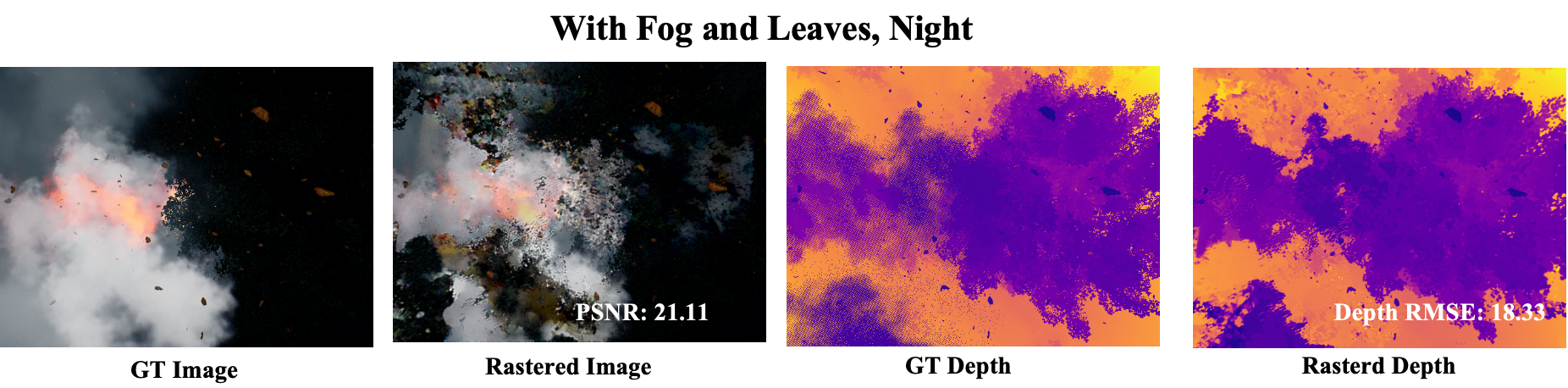}
\end{center}
% \vspace{-1em}
   \caption{Mapping results of 3DGS-based SLAM on Challenge Scenario 4 of the {\em WildfireX-SLAM} dataset.}
\label{fig:CS4}
\end{figure}

\Cref{tab:benchmarkb} presents the mapping results of Splatam across different scenes, highlighting the significant challenges that wildfire environments pose to SLAM systems, which manifest in three main aspects. (a) Dynamic non-rigid objects and repetitive textures with limited features severely challenge SLAM localization and 3DGS fitting. This performance degradation due to such objects is evident in the results for CS 1 and CS 2 in the table. As shown in \Cref{fig:CS1}, dynamic rigid objects like wildfires affect both image and depth fitting, with a particularly severe impact on depth: the Depth RMSE increases from 6.44 without fire to 14.33 with fire. (b) Illumination plays a critical role in SLAM performance. From \Cref{fig:CS2} and \Cref{tab:benchmarkb} results for CS 3, it is seen that as lighting improves from morning to noon, SLAM performance correspondingly improves. However, it is noteworthy that the PSNR metric appears to fail under very low light conditions. Specifically, \Cref{fig:CS2} shows that at night, Depth RMSE worsens compared to noon, while PSNR paradoxically improves due to loss of scene details and simpler image rendering in dark conditions, which complicates tracking. This observation suggests the need for more effective and innovative metrics tailored for nighttime scenarios. (c) Fog causes degradation in localization accuracy. Results from CS 4 in \Cref{tab:benchmarkb} and \Cref{fig:CS3} indicate that fog leads to effects similar to nighttime conditions on SLAM, with improved mapping but declined localization performance, fundamentally due to light obstruction.

\section{Conclusion}
\label{sec:conclusion}
In this paper, we introduced WildfireX-SLAM, a groundbreaking SLAM dataset designed to tackle the unique challenges of wildfire and forest environments. With its diverse and realistic data captured through both aerial and near-ground perspectives, augmented depth images, WildfireX-SLAM provides an invaluable resource for advancing SLAM technologies. Beyond the dataset itself, we offer a flexible data collection framework using Unreal Engine 5 and our modified AirSim, enabling precise control over environmental factors like wildfire conditions and weather. This work goes beyond addressing technical gaps, it lays a critical foundation for robotics and computer vision research in safety-critical domains, empowering innovations in real-time emergency response and collaborative navigation in unstructured, dynamic wild environments.

\bibliographystyle{splncs04}
\bibliography{main}

\end{document}

%% file: preamble.tex
\usepackage{pifont}
\usepackage{amssymb}
\newcommand{\xmark}{\ding{55}} % 自定义符号
\usepackage{multirow}
\usepackage{makecell}

\usepackage{hyperref}
\usepackage{cleveref}
\usepackage{booktabs}
\usepackage{xspace}
\newcommand{\eg}{e.g.\xspace}
\usepackage{caption}
\usepackage{arydshln} % 支持虚线和局部竖线

\usepackage{float}

\crefname{figure}{Fig.}{Figs.}    % 普通引用用单数和复数形式
\Crefname{figure}{Fig.}{Figs.}    % 首字母大写引用用单数和复数形式